\documentclass[]{spie}  
\usepackage{amsmath, amsfonts, graphicx, booktabs, tabularx, multirow, siunitx, xcolor}
\usepackage{hyperref}

\graphicspath{{./figures/}}

\title{Identifying disease-free chest X-ray images \\with deep transfer learning\thanks{\hspace{1em}This paper was accepted by SPIE Medical Imaging, 2019 (oral presentation). The final publication is available at \url{https://doi.org/10.1117/12.2513164}}}

\author{Ken C. L. Wong, Mehdi Moradi, Joy Wu, and Tanveer Syeda-Mahmood
\skiplinehalf
IBM Research -- Almaden Research Center, San Jose, CA, USA
}

\begin{document}
\maketitle

\begin{abstract}
Chest X-rays (CXRs) are among the most commonly used medical image modalities. They are mostly used for screening, and an indication of disease typically results in subsequent tests. As this is mostly a screening test used to rule out chest abnormalities, the requesting clinicians are often interested in whether a CXR is normal or not. A machine learning algorithm that can accurately screen out even a small proportion of the ``real normal'' exams out of all requested CXRs would be highly beneficial in reducing the workload for radiologists. In this work, we report a deep neural network trained for classifying CXRs with the goal of identifying a large number of normal (disease-free) images without risking the discharge of sick patients. We use an ImageNet-pretrained Inception-ResNet-v2 model to provide the image features, which are further used to train a model on CXRs labelled by expert radiologists. The probability threshold for classification is optimized for 100\% precision for the normal class, ensuring no sick patients are released. At this threshold we report an average recall of 50\%. This means that the proposed solution has the potential to cut in half the number of disease-free CXRs examined by radiologists, without risking the discharge of sick patients.
\end{abstract}



\section{INTRODUCTION}
\label{sec:intro}

Chest X-rays (CXRs) are among the most commonly requested medical studies in healthcare settings ranging from primary care offices, emergency departments, to intensive care units. Over 129 million CXRs were ordered in the United States alone in 2006 \cite{Journal:Mettler:Radiology2009}, contributing to a significant portion of radiologists’ daily workload. As this is mostly a screening test used to rule out chest abnormalities, the requesting clinicians are often interested in whether a CXR is normal or not.

With the large CXR datasets, such as the ChestX-ray14 dataset from the National Institutes of Health (NIH) \cite{Conference:Wang:CVPR2017} and the MIMIC-CXR dataset from the MIT Laboratory for Computational Physiology \cite{Journal:Johnson:arXiv2019}, become available, different deep learning frameworks have been proposed for abnormal findings classification from CXRs. In \cite{Conference:Wang:CVPR2017}, a unified deep convolutional neural network framework was proposed, which allows the use of different ImageNet-pretrained models for abnormal findings classification and localization, and the results on the eight abnormal findings of the ChestX-ray14 dataset were reported. In \cite{Journal:Rajpurkar:arXiv2017}, a 121-layer DenseNet was trained on the ChestX-ray14 dataset for findings classification and localization, and the results on the 14 abnormal findings were reported.

Although the results of the existing frameworks are promising, accurate classification of multiple abnormal findings is still a very difficult task. In fact, instead of classifying multiple findings, a machine learning algorithm that can accurately screen out even a small proportion of the ``real normal'' exams out of all requested CXRs would be highly beneficial in reducing the workload for radiologists and provide meaningful clinical feedback to the requesting clinicians. To have a high confidence that no abnormalities exist among the CXRs identified as normal, the primary goal of this machine learning algorithm would be to penalize any false positives for the normal class heavily. In other words, it would need to demonstrate a high precision for predicting normal.

In consequence, here we report a machine learning solution for classifying a given CXR as normal or abnormal, with a very high precision for the normal class. This solution uses a deep neural network acting as a binary classifier. We show that this classifier can correctly identify nearly 50\% of the normal (disease-free) images without mislabeling any images showing diseases.

\begin{table*}[t]
\caption{Abnormal findings in the ``Abnormal'' class.}
\label{table:abnormal}
\medskip
\centering
\begin{tabularx}{\linewidth}{XX}
\toprule
Alveolar opacity & Mass plus nodule\\
Atelectasis & Pleural effusion \\
Enlarged cardiac silhouette & Pleural mass plus thickening \\
Hernia & Pneumothorax \\
Hyperaeration & Vascular redistribution \\
Increased reticular markings & \\
\bottomrule
\medskip
\end{tabularx}
\end{table*}

\begin{figure}[t]
    \medskip
    \centering
    \begin{minipage}[b]{0.32\linewidth}
      \centering
      \includegraphics[width=1\linewidth]{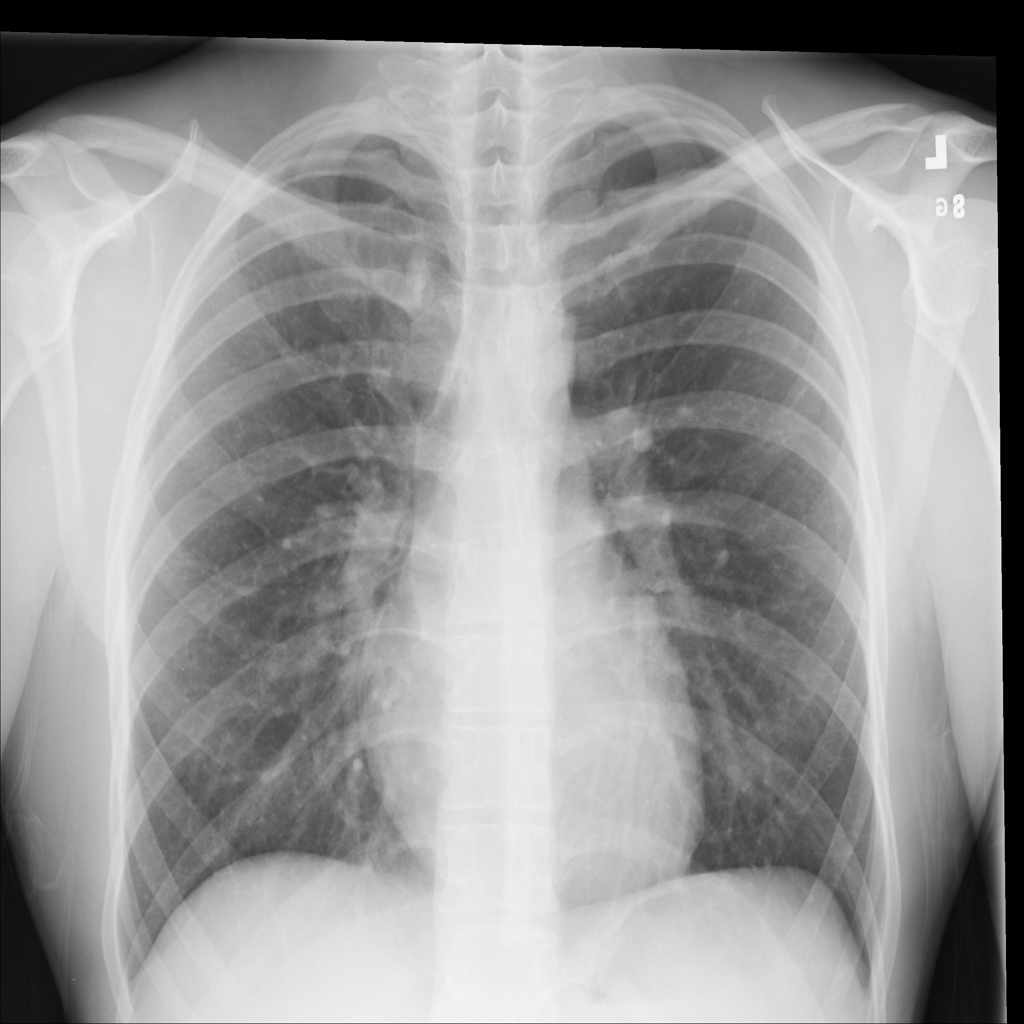}
      \centering{Normal}
    \end{minipage}
    \begin{minipage}[b]{0.32\linewidth}
      \centering
      \includegraphics[width=1\linewidth]{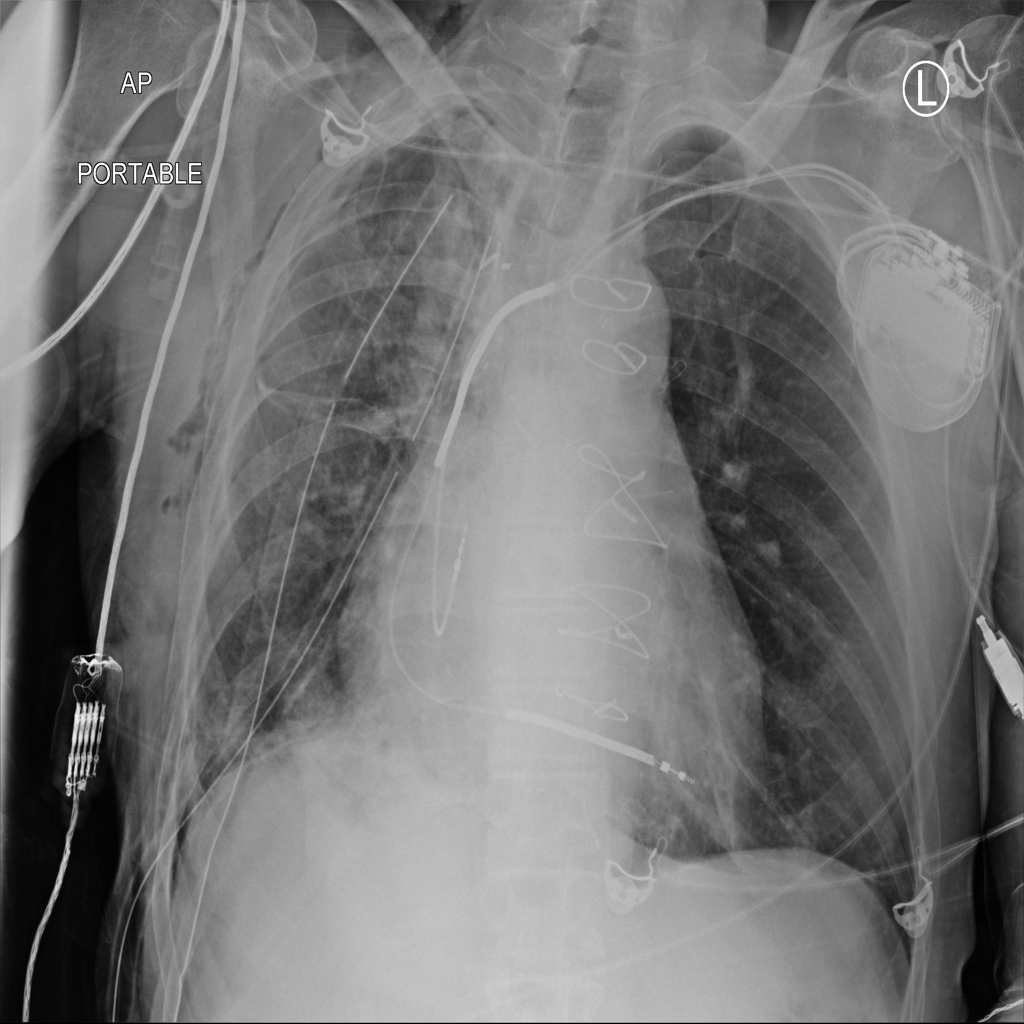}
      \centering{Pneumothorax and hyperaeration}
    \end{minipage}
    \begin{minipage}[b]{0.32\linewidth}
      \centering
      \includegraphics[width=1\linewidth]{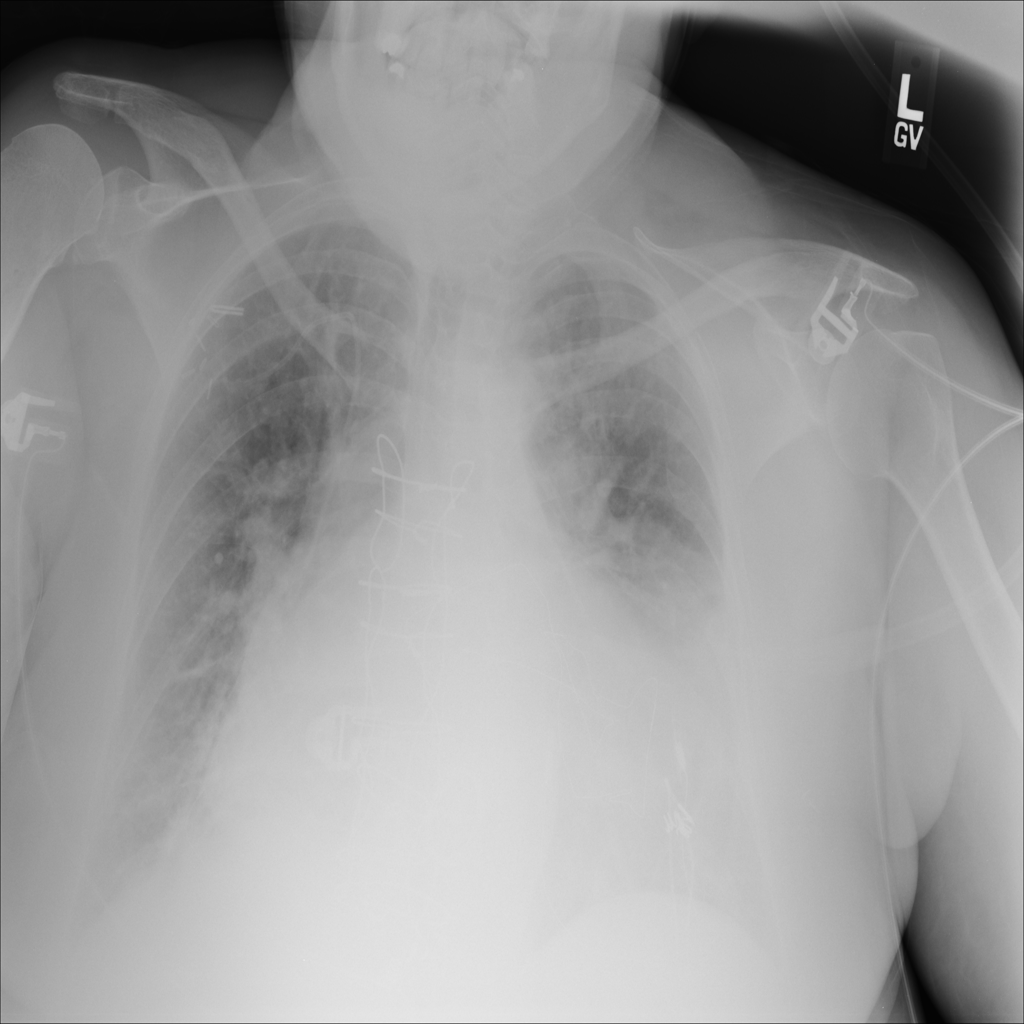}
      \centering{Enlarged cardiac silhouette}
    \end{minipage}
    \\
    \medskip
    \begin{minipage}[b]{0.32\linewidth}
      \centering
      \includegraphics[width=1\linewidth]{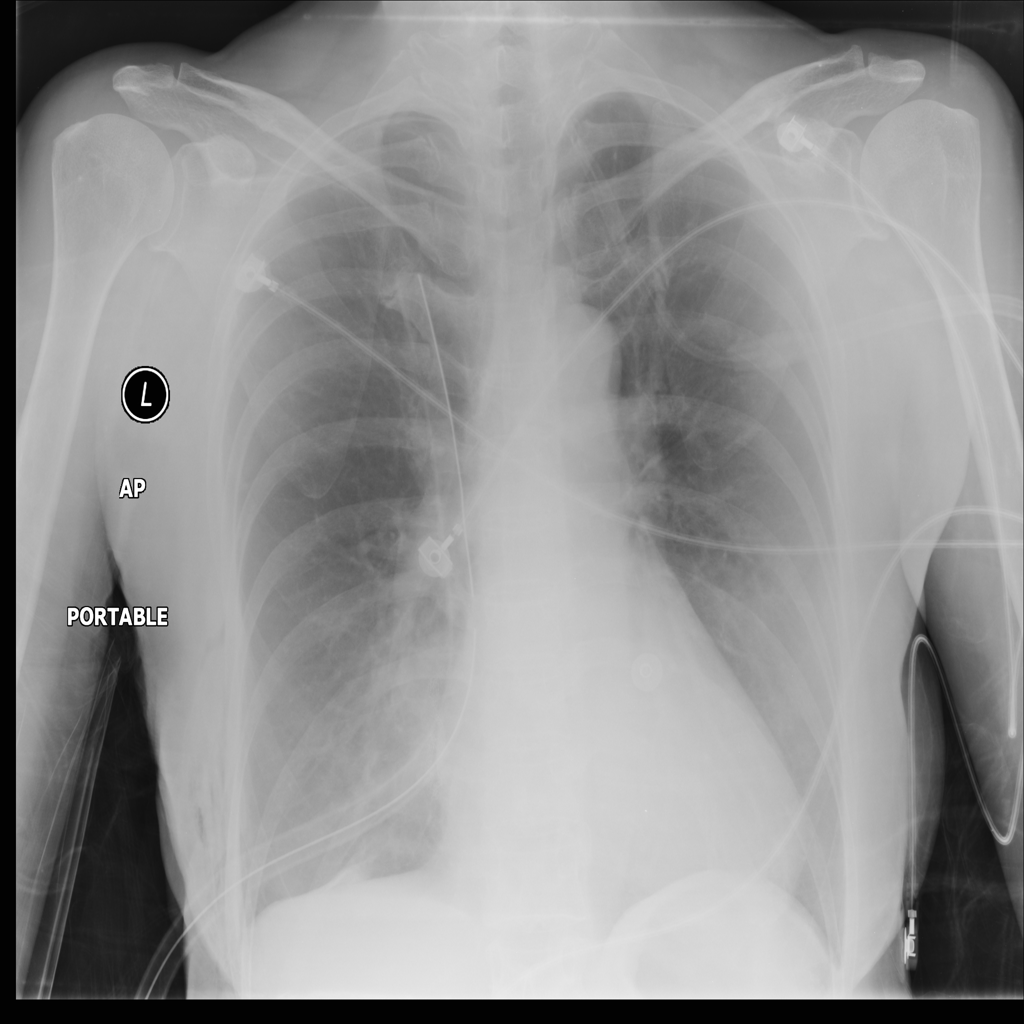}
      \centering{Hyperaeration}
    \end{minipage}
    \begin{minipage}[b]{0.32\linewidth}
      \centering
      \includegraphics[width=1\linewidth]{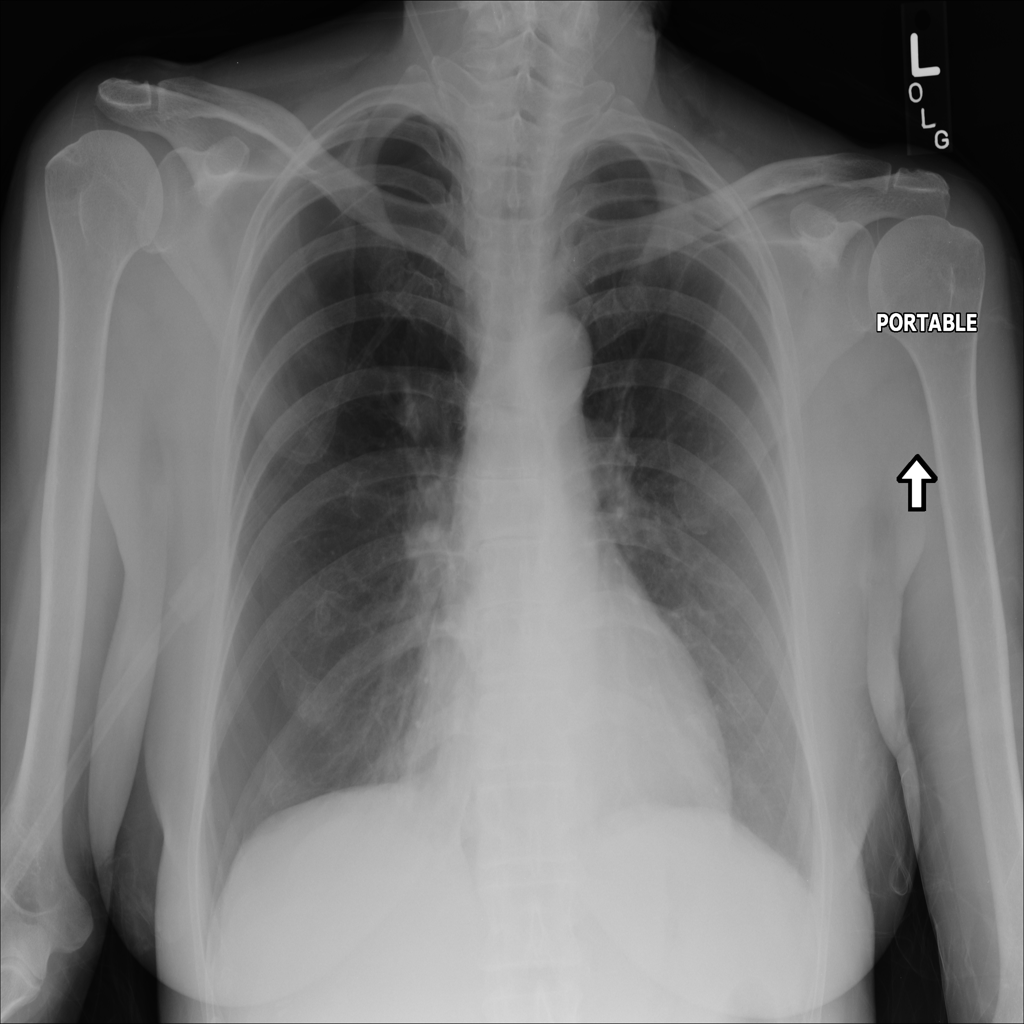}
      \centering{Pneumothorax}
    \end{minipage}
    \begin{minipage}[b]{0.32\linewidth}
      \centering
      \includegraphics[width=1\linewidth]{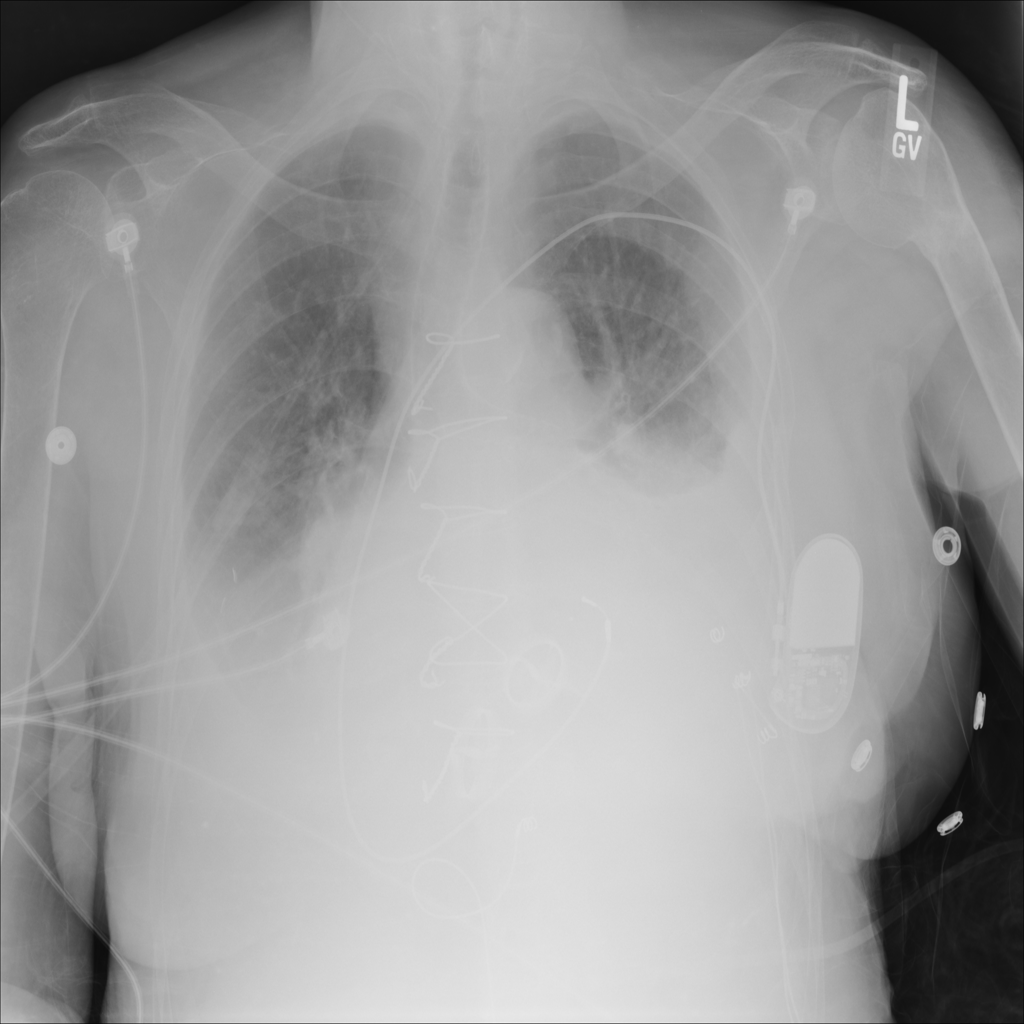}
      \centering{Pleural effusion}
    \end{minipage}
    \\
    \medskip
    \caption{Examples of normal and abnormal CXRs.}
    \label{fig:examples}
\end{figure}

\section{METHOD}

\subsection{Data}

The images used in this study are from the NIH chest X-ray dataset, ChestX-ray14 \cite{Misc:NIHChestXray2017,Conference:Wang:CVPR2017}. The dataset was extracted from the clinical PACS database at the NIH Clinical Center and consists of around 60\% of all frontal CXRs in the hospital. The images were annotated by NIH through a variety of natural language processing techniques applied on the associated radiological reports. Although this dataset with over 100,000 images from more than 30,000 unique patients is very useful for training disease classifiers, the dataset only indicates those images that do not contain any of their target findings as ``No Finding'' and the ``Normal'' class does not exist. In fact, the definition of normal can be ambiguous depending on the type of CXRs. For example, an AP CXR may show tubes and lines but is otherwise normal in anatomic findings. Therefore, we created a freshly labeled dataset from the original NIH data in which we asked clinicians to mark the normal images. Experienced radiologists examined a subset consisting of 3000 AP images from the NIH dataset and identified 1300 images with no pathologic finding. An additional 1917 AP images from the NIH dataset with disease labels were also examined by experienced contracting radiologists and confirmed to display one or more abnormal findings. This gave us a total of 3217 images used in this work. The types of abnormal findings in the ``Abnormal'' class are listed in Table \ref{table:abnormal} and examples of normal and abnormal images are shown in Figure \ref{fig:examples}.

\subsection{Network architecture}

\begin{figure}[t]
    \centering
    \begin{minipage}[b]{1\linewidth}
      \centering
      \includegraphics[width=1\linewidth]{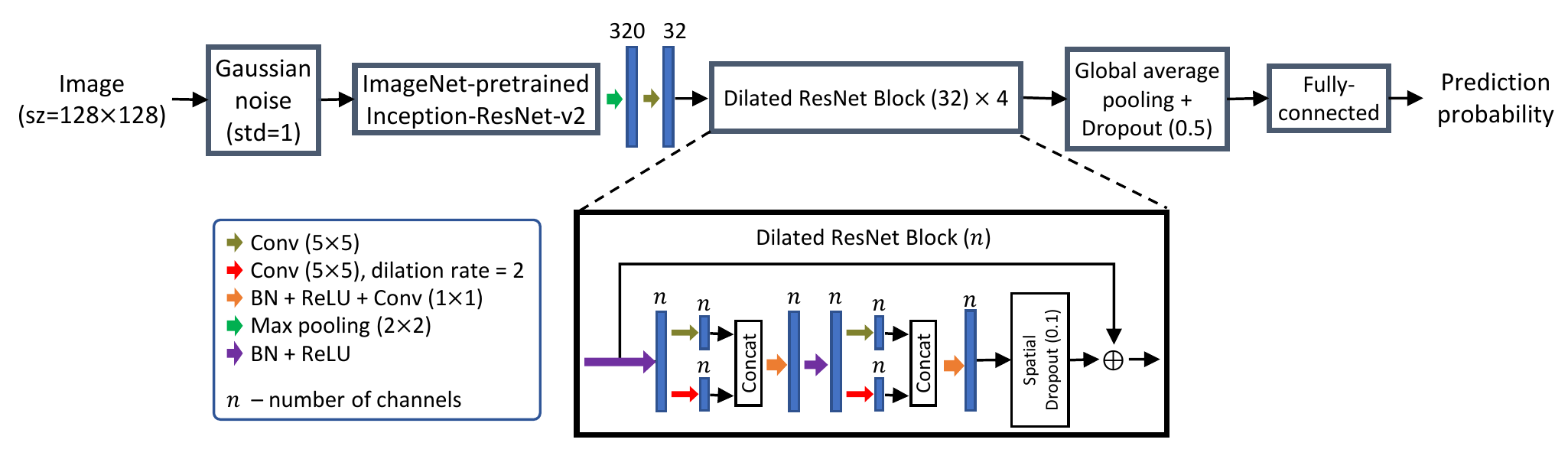}
    \end{minipage}
    \caption{Network architecture. The ImageNet-pretrained Inception-ResNet-v2 model is used to provide image features for further training. The Keras implementation of Inception-ResNet-v2 is used and the 320 feature channels from layer ``mixed\_5b'' are used to provide the ImageNet-pretrained features. The Dilated ResNet Block is proposed to obtain multi-scale features with fast convergence. Gaussian noise, spatial dropout, and standard dropout are used to reduce overfitting and improve model generalization.}
    \label{fig:architecture}
\end{figure}

Given the relatively small number of images, transfer learning is required to train a convolutional neural network (CNN) with reasonable performance. The network architecture is shown in Figure \ref{fig:architecture}. The ImageNet-pretrained Inception-ResNet-v2 model \cite{Conference:Szegedy:AAAI2017} is used with our proposed architecture. Inception-ResNet-v2 combines the advantages of Inception networks and residual connections \cite{Conference:He:CVPR2016,Conference:He:ECCV2016} to achieve state-of-the-art accuracy on the ILSVRC image classification benchmark. As the features from the deeper layers of a pretrained model may be too problem-specific, the relatively low-level features are used. The Keras implementation of Inception-ResNet-v2 is used and the 320 feature channels from layer ''mixed\_5b'' are used to provide the ImageNet-pretrained features obtained by 12 convolutional layers \cite{Misc:Chollet:Keras2015}.

To learn the problem-specific features from the ImageNet-pretrained features, here we propose the Dilated ResNet Block (Figure \ref{fig:architecture}). Each block consists of 5$\times$5 convolutions and 5$\times$5 dilated convolutions with dilation rate = 2 in parallel, whose outputs are concatenated. Dilated convolution is used to obtain multi-scale features without increasing computational complexity \cite{Journal:Yu:arXiv2015}. Like the ResNet block, skip connection is used for better convergence, and identity mapping is used with after-addition activation to facilitate more direct information propagation \cite{Conference:He:ECCV2016}. The spatial dropout, which drops entire feature maps instead of individual elements, is also used in each block for more effective overfitting reduction \cite{Conference:Tompson:CVPR2015}. Four Dilated ResNet Blocks are cascaded, followed by global average pooling and a final fully-connected layer with the sigmoid function to produce the classification probabilities. Standard dropout is used after global average pooling to further reduce overfitting. A Gaussian noise layer with the standard deviation as one is also used to improve model generalization.

\subsection{Training strategy}

Images were normalized by contrast limited adaptive histogram equalization to enhance contrast and were resized to 128$\times$128 for faster training. Image augmentation was performed with rotation ($\pm$\ang{10}), shifting ($\pm$10\%), and scaling ([0.95, 1.05]) to learn invariant features and reduce overfitting, and each image had an 80\% chance to be transformed during training. The binary cross-entropy was used to compute the loss function, and we define ``Normal'' as the positive class (class 1) and ``Abnormal'' as the negative class (class 0) as our target group is normal. The optimizer Adam for first-order gradient-based optimization of stochastic objective functions was used for fast convergence \cite{Journal:Kingma:arXiv2014}, with the learning rate as 10$^{-4}$. Each training had 50 epochs with a batch size of 400. Five-fold cross-validations, in terms of unique patients, were performed. The Python deep learning library Keras \cite{Misc:Chollet:Keras2015} with the TensorFlow backend \cite{Conference:Abadi:OSDI2016} was used for the implementation, and an NVIDIA Tesla P100 with 16 GB of memory was used.

\begin{figure}[t]
    \medskip
    \centering
    \begin{minipage}[b]{0.49\linewidth}
      \centering
      \includegraphics[width=1\linewidth]{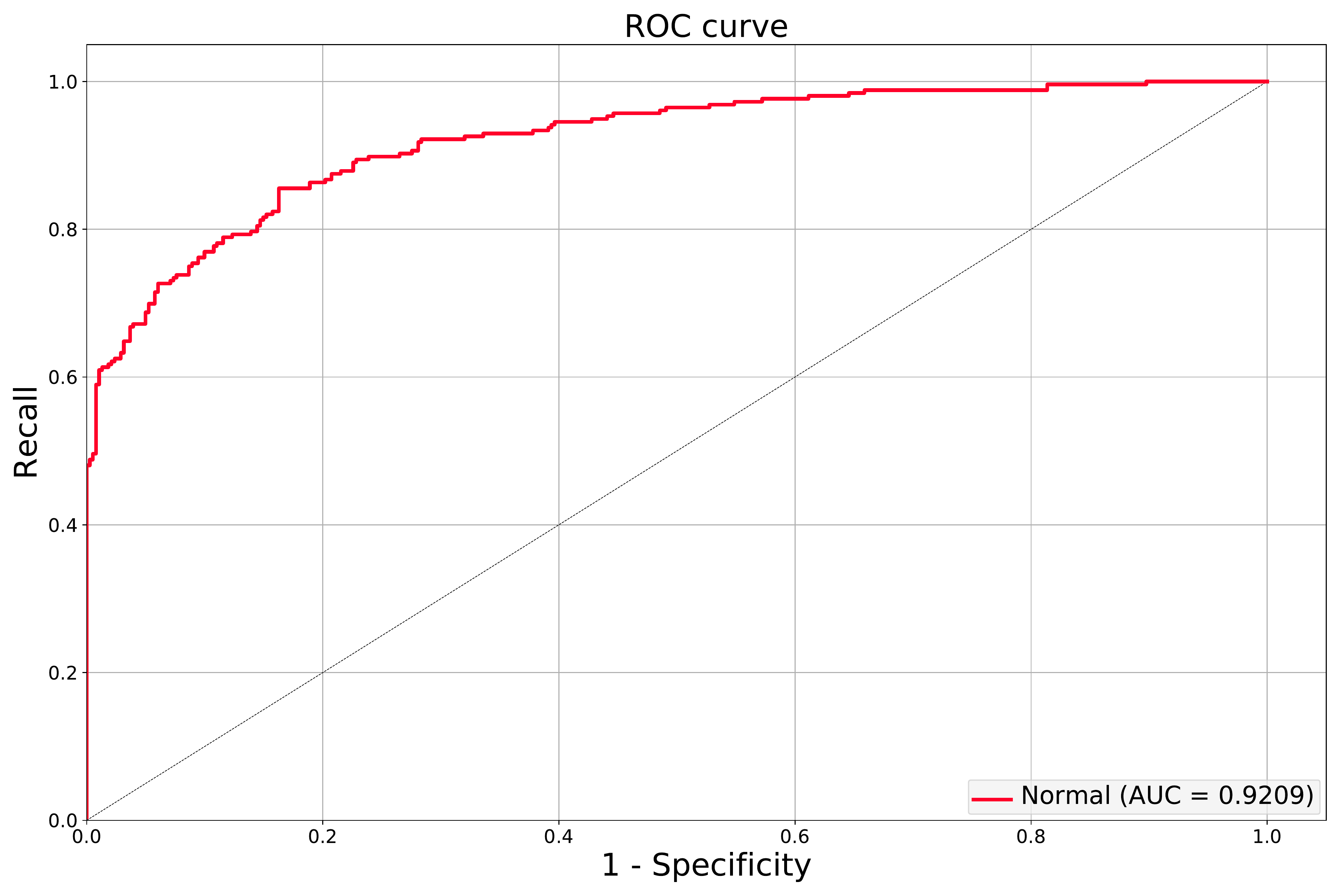}
    \end{minipage}
    \begin{minipage}[b]{0.49\linewidth}
      \centering
      \includegraphics[width=1\linewidth]{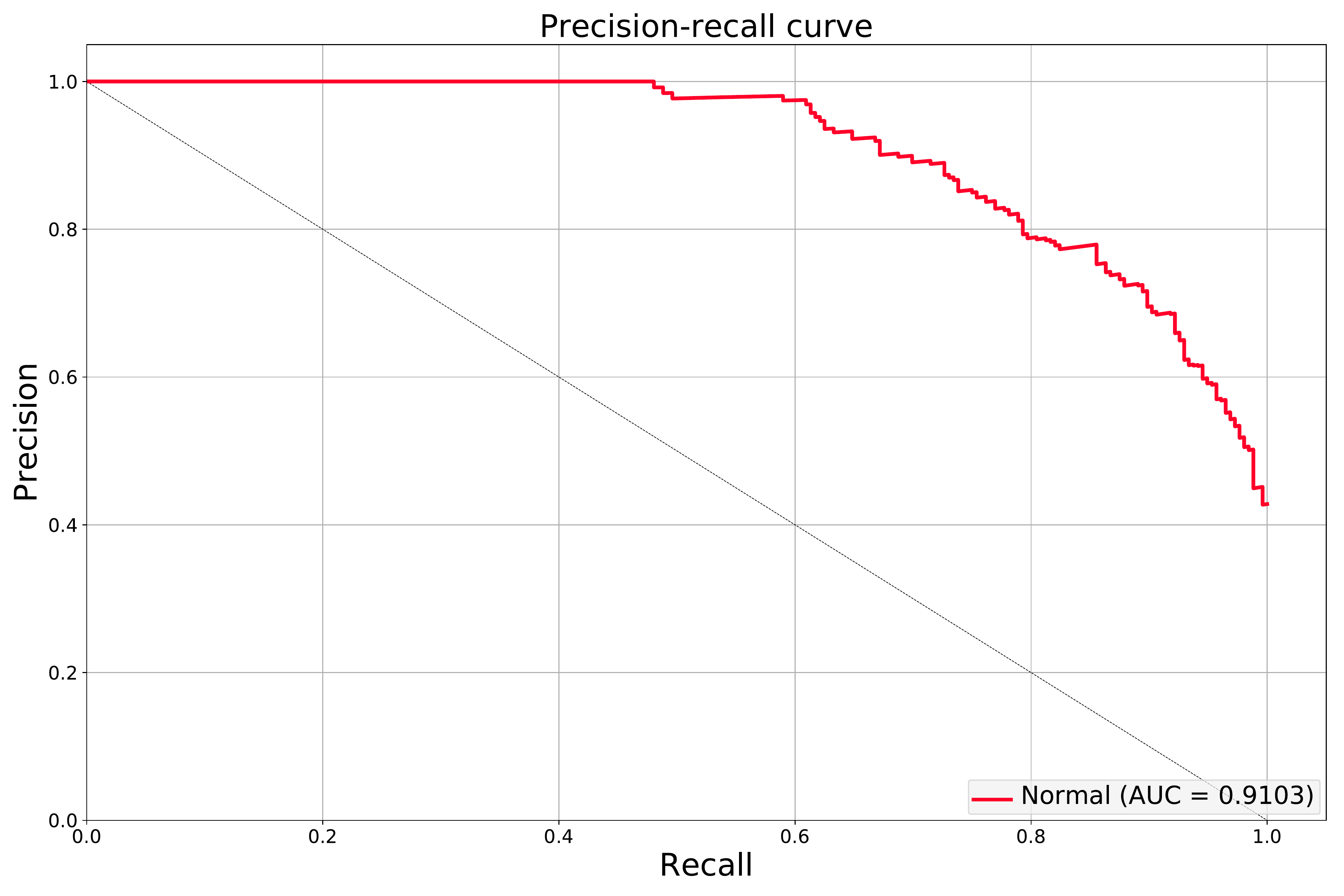}
    \end{minipage}
    \caption{Classification performance of a validation. Left: ROC curve (AUC = 0.92). Right: precision-recall curve (AUC = 0.91).}
    \label{fig:plots}
    \medskip
\end{figure}

\begin{table*}[t]
\caption{Results of five-fold cross-validations. The probability thresholds were computed by optimizing the precision scores, and the corresponding true positives (TP), false positives (FP), true negatives (TN), false negatives (FN), recalls, and specificities are shown.}
\label{table:crossvalidations}
\smallskip
\centering
\begin{tabularx}{\linewidth}{XXXXXXXXXX}
\toprule
Test & ROC AUC & PR AUC & Threshold & TP & FP & TN & FN & Recall & Specificity \\
\midrule
0 & 0.92 & 0.91 & 0.99 & 121 & 0 & 381 & 135 & 0.47 & 1.00 \\
1 & 0.89 & 0.88 & 0.99 & 122 & 0 & 390 & 146 & 0.46 & 1.00 \\
2 & 0.93 & 0.91 & 0.98 & 138 & 0 & 378 & 117 & 0.54 & 1.00 \\
3 & 0.88 & 0.88 & 0.99 & 119 & 0 & 387 & 141 & 0.46 & 1.00 \\
4 & 0.90 & 0.90 & 0.81 & 146 & 0 & 381 & 115 & 0.56 & 1.00 \\
\midrule
Min & 0.88 & 0.88 & 0.81 & 119 & 0 & 378 & 115 & 0.46 & 1.00 \\
Max & 0.93 & 0.91 & 0.99 & 146 & 0 & 390 & 146 & 0.56 & 1.00 \\
Average & 0.90 & 0.90 & 0.95 & 129.20 & 0.00 & 383.40 & 130.80 & 0.50 & 1.00 \\
Std & 0.02 & 0.02 & 0.08 & 12.07 & 0.00 & 4.93 & 14.08 & 0.05 & 0.00 \\
\bottomrule
\end{tabularx}
\end{table*}

\section{Results and Discussion}

As we define ``Normal'' as positive and ``Abnormal'' as negative, we want to minimize false positives (sick patients diagnosed as normal) as delayed treatment can have severe consequences. On the other hand, we want to reduce the workload of radiologists, thus a reasonably high percentage of true positives (normal patients diagnosed as normal) which can be removed from further reading is desired. This means that a very high precision score and a reasonably high recall score are required, which are defined as:
\begin{gather}
\label{eq:recall_precision}
    \mathrm{Recall} = \frac{TP}{TP + FN}, \ \ \ \ \mathrm{Precision} = \frac{TP}{TP + FP}
\end{gather}
with $TP$, $FN$, and $FP$ as true positives, false negatives, and false positives, respectively.

Figure \ref{fig:plots} shows the results of a validation. Apart from the receiver operating characteristic (ROC) curve, we also show the precision-recall curve for a more complete evaluation of the performance. The areas under curve (AUC) of both curves are larger than 90\%. When the probability threshold increases, the false positive rate (1 - specificity) and the precision score remain as 0 and 1, respectively, until the recall score is near 0.5. As ``Normal'' is the positive class, this means that we can maintain low false positives which is important to minimize the risks of sick patients being wrongly discharged.

Table \ref{table:crossvalidations} shows the results of the five-fold cross-validations. The performances of all validations were very consistent, with both the average AUC of the ROC and precision-recall curves as 90\% and the standard deviations as 2\%. To show the best possible performance in terms of minimizing false positives, we obtained the probability thresholds by optimizing the precision scores using the global optimization algorithm DIRECT (DIviding-RECTangles) \cite{Journal:Gablonsky:JGO2001}. DIRECT is a pattern search algorithm that balances local and global search to efficiently find a globally optimized value, and it is designed to completely explore the
searching space even after multiple local minima are identified. The optimal thresholds and the corresponding metrics are shown in Table \ref{table:crossvalidations}. The thresholds were very close to one so that only images which were almost certain to be normal were classified as normal and thus the false positives can be minimized. The false positives were zeros in all validations, and the average recall score was 50\%. This means that the clinicians could reduce 50\% workload from the normal patients with minimal risks of discharging sick patients.

\section{Conclusion}

We introduce a deep learning framework for normal/abnormal classification. Using features provided by the ImageNet-pretrained Inception-ResNet-v2 model and the proposed Dilated ResNet Block which comprises the advantages of dilated convolution and residual connection, an efficient CNN model that utilizes multi-scale features can be trained. The AUC of the ROC and precision-recall curves of the five-fold cross-validations show promising performance. Using the probability thresholds obtained by optimizing the precision scores with global optimization, this framework shows potential to reduce radiologists' workload while minimizing the risks of discharging sick patients.

\bibliography{Ref}   
\bibliographystyle{spiebib}   

\end{document}